\documentclass[utf8]{FrontiersinHarvard} 
\usepackage{url,hyperref,lineno,microtype,subcaption} 
\usepackage[onehalfspacing]{setspace}
\usepackage{gensymb}
\usepackage{comment}
\usepackage[normalem]{ulem}
\usepackage{cleveref}
\usepackage{tabularx}
\usepackage{tablestyles}
\usepackage{natbib}
\usepackage{bibunits}
\defaultbibliographystyle{IEEEtranN} 
\defaultbibliography{main}

\def\keyFont{\fontsize{8}{11}\helveticabold }
\def\firstAuthorLast{Lessons from a Space Lab} 
\def\Authors{Leo PAULY, Michele Lynn JAMROZIK, Miguel ORTIZ DEL CASTILLO, Olivia BORGUE, Inder Pal SINGH, Mohatashem Reyaz MAKHDOOMI, Olga-Orsalia CHRISTIDI-LOUMPASEFSKI, Vincent GAUDILLIERE, Carol MARTINEZ, Arunkumar RATHINAM, Andreas HEIN, Miguel OLIVARES-MENDEZ, Djamila AOUADA}
\def\Address{Interdisciplinary Centre for Security, Reliability and Trust (SnT), University of Luxembourg, Luxembourg}
\def\corrAuthor{}
\def\corrEmail{}

\begin{document} 
\onecolumn
\firstpage{1}
\title{Lessons from a Space Lab - An Image Acquisition Perspective}
\author[\firstAuthorLast ]{\Authors} 
\address{} 
\correspondance{leo.pauly@uni.lu} 
\extraAuth{}%
\maketitle

\begin{bibunit}
\begin{abstract}
\section{}
The use of Deep Learning (DL) algorithms has improved the performance of vision-based space applications in recent years. However, generating large amounts of annotated data for training these DL algorithms has proven challenging. While synthetically generated images can be used, the DL models trained on synthetic data are often susceptible to performance degradation, when tested in real-world environments.  
In this context, the Interdisciplinary Center of Security, Reliability
and Trust (SnT) at the University of Luxembourg has developed the {`SnT Zero-G Lab'}, for training and validating vision-based space algorithms in conditions emulating real-world space environments. 
An important aspect of the SnT Zero-G Lab development was the equipment selection. From the lessons learned during the lab development, this article presents a systematic approach combining market survey and experimental analyses for equipment selection. In particular, the article focus on the image acquisition equipment in a space lab: background materials, cameras and illumination lamps. The results from the experiment analyses show that the market survey complimented by experimental analyses is required for effective equipment selection in a space lab development project. 
 
\tiny
 \keyFont{ \section{Keywords:} Space Lab, Image Acquisition, Background Analysis, Lighting, Exposure Settings} 
\end{abstract}


\section{Introduction}

In the last few years, Deep Learning (DL) techniques have been proven successful in vision-based space applications such as satellite pose estimation
\citep{Dung_2021_CVPR} \citep{kisantal2020} and spacecraft navigation \citep{song2022deep}. However, DL models require a vast amount of annotated data to learn data patterns and achieve a high performance. Nevertheless, due to the difficulties of obtaining large real space datasets with correct labels, robust  space-related DL-solutions are currently missing. For that reason, previous research has mostly relied on synthetic data for training DL models~\citep{Musallam_2021_ICCV}\citep{kisantal2020}\citep{proenca2019}. However, while synthetic images are easy to generate and annotate for training DL-based solutions, they are prone to performance degradation when the model is tested in a real world environment, as DL solutions tend to over-fit the features from the synthetic domain \citep{peng2017visda}. This phenomenon is known as the \textit{Domain Gap} problem \citep{zhang2022micro}\citep{chen2022sim}.


To address the domain gap problem, several research institutions around the world, such as European Proximity Operations Simulator (EPOS) \citep{EPOS2011} and Autonomous Spacecraft Testing of Robotic Operations in Space (ASTROS) \citep{ASTROS2018}, are developing laboratory facilities for mimicking space conditions with the objective of obtaining more reliable datasets. Research suggests that real world space-like image datasets (Spacecraft PosE Estimation Dataset+ (SPEED+) ~\citep{park2021}, SPAcecraft Recognition leveraging Knowledge of space environment 2022 (SPARK-2022) \citep{rathinam_arunkumar_2022_6599762}) collected in these facilities can be used to train and evaluate the robustness of vision-based space algorithms, mitigate the domain gap, and provide higher confidence on the performance of the DL models when deployed in space \citep{park2021}. 
However, the construction of such facility entails a plethora of uncertainties as it is not a standardized nor well-documented process. Consequently, research centers undertaking this endeavour face many challenges, including a lack of support in the form of guides, manuals, or templates \citep{beierle2019high}\citep{boge2011using}\citep{cassinis2022ground}. As any other development project, the major drawback from these uncertainties is the increased probability of  cost overruns, project delays, and even project failure \citep{raz2002risk}. 

In 2019, the Interdisciplinary Center of Security, Reliability and Trust (SnT) at University of Luxembourg undertook the project of developing the `SnT Zero-G Lab', a facility for mimicking space environment and simulate rendezvous related processes. 
During the development, SnT Zero-G Lab had faced cost and schedule challenges related to the lack of literature documenting the development of such facilities.
This article belongs to a series of articles that SnT is producing with the objective of bringing forward the lessons learned during the development of the SnT Zero-G Lab, and supporting research institutions around the world in developing their own space facilities.

An important aspect of building a space facility like SnT Zero-G Lab is the equipment selection. The selection of the right equipment to emulate a space-like environment and capture images of acceptable quality level in such conditions is crucial. Hence it is important to have details on the available options in the market, selection metrics to be used, and experimental analysis methods for equipment comparison that would support purchase decisions. In this context, the goal of this article is to provide a systematic approach to support equipment selection and decision making when developing a space lab for vision-based applications. In particular, the article focuses on the equipment required in the image acquisition process, the laboratory background materials, cameras and illumination lamps. 

The contributions are summarised below:

\begin{itemize}
\item A detailed survey of market available choices for image acquisition equipment. Background materials, cameras and illumination lamps were surveyed and compared based on selection metrics. The selection metrics were chosen based on their relevance to the image capturing process (for example, focal length and shutter speed) as well as the lab development project objectives (for example, cost and size). 
\item Experimental analyses for equipment comparison for background materials and cameras. 
\item Together, the survey and experimental analyses provide a systematic approach for equipment selection in the development of similar space labs. The presented framework can be extended to include more products when available in future and make a selection based on different project objectives such as budget constraints and intended applications.   
\end{itemize}

The remaining of the paper is organised as follows. First, a literature review of existing space facilities with a focus on image acquisition components is summarized in \Cref{sec:literature}. Then, a market survey of laboratory background materials, cameras and illumination equipment is presented in \Cref{sec:survey_materials}. 
The survey is then complemented with different experiments to analyse the suitability and performance of commercially available equipment. In \Cref{sec:labsetup_datacollection}, the laboratory setup is described, and in \Cref{sec:backgroundtest} and \Cref{sec:exposuretest} experimental analyses of laboratory backgrounds and cameras are presented. Section \ref{sec:discussion} discusses the results and Section \ref{sec:conclusion} concludes the paper.

\section{Related Work}
\label{sec:literature}

Several facilities providing testbeds for training and validating vision-based space applications exist at different research institutions around the world. In this section, a review of these facilities with a focus on image acquisition components (background materials, cameras and illumination) is presented.

\subsection{TRON, USA}
The Robotic Testbed for Rendezvous and Optical Navigation (TRON) facility at Stanford's Space Rendezvous Laboratory (SLAB) is the first of its kind developed for testing machine learning based space-borne optical navigation algorithms \citep{TRON2021}. The TRON facility can accurately reproduce a wide range of lighting scenarios representative of the space environment. To mimic the diffused light of Earth's albedo, ten light boxes are installed around the walls. Each light box consists of a diffuser plate covered with hundreds of Light Emitting Diodes (LEDs) organized in strips and can adjust their colour and intensity. The light boxes are calibrated to produce radiance across the diffuser plates that are as uniform as possible and compatible with Earth's albedo in Low Earth Orbits (LEO). A metal halide arc lamp is also used at the facility to simulate direct sunlight. As the background material, light-absorbing black commando curtains are placed over all ambient light sources, including the windows and the deactivated light boxes, to enhance the impact of diffused and direct light \citep{park2021}. 

\subsection{GRALS, Netherlands}
The GNC Rendezvous, Approach and Landing Simulator (GRALS) testbed is situated in the  Orbital Robotics and GNC Laboratory (ORGL) at the European Space Research and Technology Centre (ESTEC) \citep{GRALS-ESTEC2022}. 
A Prosilica GC2450 camera mounted on a KUKA robotic arm is used for capturing images at the facility. To recreate a realistic space environment from an illumination standpoint, a movable lamp is mounted on an UR-5 robot and directed towards the target mockup during image acquisition. The lamp is a dimmable, uniform, and collimated light source with a spectral response close to 6000K and exclusive optical lens which provide high uniformity (±5\%) shadow-free backlight illumination. Besides, black background curtains are placed around the robots’ workspace in order to mask most of the background noise, such as unwanted reflections from the robots’ rails.

\subsection{EPOS, Germany}

The European Proximity Operations Simulator (EPOS) test facility was developed by the German Aerospace Center 
(DLR) to study rendezvous and docking scenarios  \citep{EPOS2011}. Two different types of cameras are used at the facility: two Charge Couple Device (CCD) Prosilica GC-655M cameras for capturing intensity images (in the visible spectrum) and two Photonic Mixer
Device (PMD) cameras (PMDtec Camcube 3.0 and Bluetechnix Argos3D-IRS1020 DLR Prototype) for capturing depth images. For simulating realistic illumination, an ARRI Max 18/12 theater spotlight is used \citep{EPOS2017}. This daylight spotlight is equipped with a hydrargyrum medium-arc iodide (HMI) light source and can generate a spectrally realistic irradiation, resembling that in the Earth’s orbit around the Sun. The spotlight is mounted on a 2-DOF yoke that is electrically steerable for easy fine-tuning of the  illumination direction.
To capture images of the target satellite with a space-realistic background, black curtains are used as background and the robotic arm carrying the satellite mockup is wrapped with black molton material.  

\subsection{ASTROS, USA}
The Autonomous Spacecraft Testing of Robotic Operations in Space (ASTROS) platform at the Georgia Institute of Technology supports experiments in vision-based autonomous rendezvous and docking, with a focus on on-orbit servicing of spacecrafts \citep{ASTROS2018}. 
The platform is equipped with a monocular PointGrey Flea3 camera to capture images (and videos) in different resolutions. The facility is capable of producing realistic images by various configurations of lighting  and dark environments and can replicate the harsh contrasts of imaging highly reflective surfaces against a dark background as seen in space. However, the technical details of the equipment used for creating these different illumination conditions are not available publicly. An overhead projector on the ceiling projects virtual images from Earth orbit against a projection screen on the wall, \citep{ASTROS2014}, which serves as the background for images captured.

\subsection{ORION, USA}
The Florida Institute of Technology developed the Orbital Robotic Interaction, On-orbit servicing, and Navigation (ORION) laboratory to test spacecraft GNC systems for proximity manoeuvres and autonomous or telerobotic capture \citep{ORION2016}. The ORION simulator uses the commercial-off-the-shelf Litepanels Hilio D12 LED panel to generate a light source sufficiently bright to exceed the dynamic range of common optical sensors while providing a narrow beam angle. The panel generates light with a color temperature of 5600 K (daylight balanced) with 350 W of power. The intensity can be continuously dimmed from 100\% to 0\%, and the beam angle can be varied between 10° and 60° using lens inserts. The light can  be used not only to simulate solar illumination, but also the weaker and diffused Earth's albedo. The background walls, floor, and ceiling of the testbed are painted a low-reflectivity black paint and all windows are covered with black-out blinds to fully control the lighting conditions and to reproduce orbital conditions.

\subsection{INVERITAS, Germany}

Innovative Technologies for Relative Navigation and Capture of Mobile Autonomous Systems (INVERITAS) facility at the Robotics Innovation Center of the German Research Center for Artificial Intelligence (RIC DFKI), designed and constructed under the INVERITAS project, models rendezvous and capture manoeuvres between a client satellite and a servicer satellite in Earth orbit \citep{INVERITAS2015}. The facility is equipped with six mobile spotlights to reproduce space-like illumination conditions. Each spotlight is motorized, allowing pan and tilt rotations and the field of view can be varied between 12° and 30°. The spotlights can be moved up and down from 1-6m. The 575W gas discharge lamps used at the facility deliver a 6000K light, with maximal intensity of 14500 Lux at 10m distance and a 12° field of view. Special light absorbing paints are used on the background walls, ceiling, and on all visible components of the system providing a space-like non-reflective background.

The reviewed literature about vision-based laboratories for space applications presents a summary of the facilities and the image acquisition  equipment used. However, details about the different materials and equipment options considered during laboratory development are missing. The logic behind their equipment selection is missing from the literature.
Moreover, the fact that every reviewed facility has been developed with different materials and equipment suggests that many commercially available alternatives can be considered to attain similar or equivalent objectives. 
In addition, each lab development project will probably have different scope and limitations in terms of budget, space available, intended applications, and other resources.   
Hence it is interesting to understand the benefits and drawbacks of the different market-available equipment options in terms of cost, ability to emulate space-like environment, quality of images captured and other relevant factors.
The following section presents a detailed market survey with comparison metrics for commercially available materials and equipment for a vision-based space applications lab. 

\section{Survey of Materials and Equipment}
\label{sec:survey_materials}
\begin{figure}[!t]
	\centering
	\includegraphics[scale=.5]{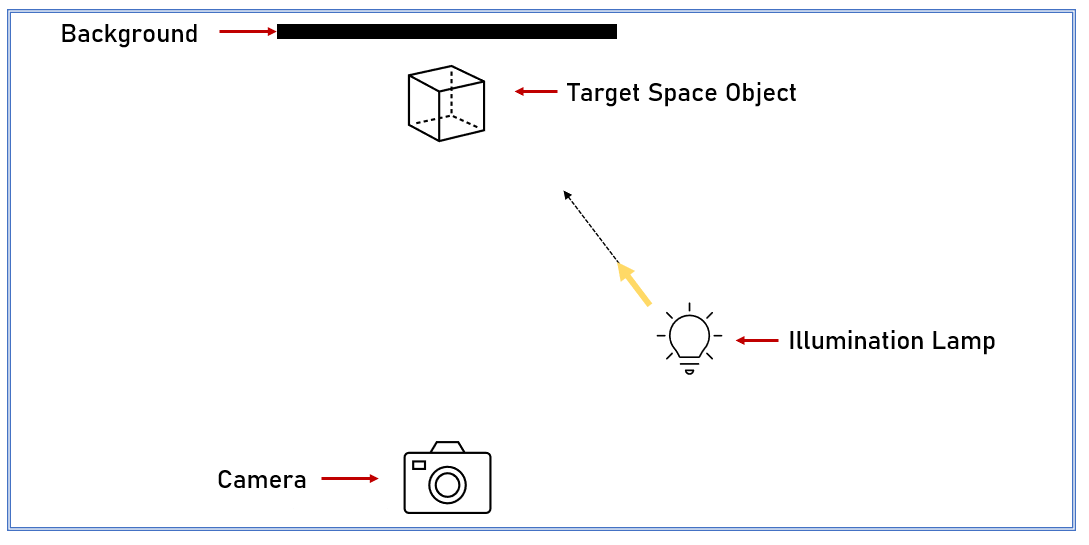}
	\caption{Image acquisition process in a space lab facility illustrated schematically.}
	\label{fig:lab_schematic}
\end{figure}

This section presents a review of commercially available equipment required for image acquisition at a space lab, as illustrated schematically in \Cref{fig:lab_schematic}. In this review,  background materials (\Cref{sec:bg_survey}), cameras (\Cref{sec:camera_survey}) and illumination lamps (\Cref{sec:lamps_survey}) to recreate high-fidelity space conditions, are included. The reference links for each item reviewed are given in supplementary material Section A. 
 
\subsection{Background materials}
\label{sec:bg_survey}

The surveyed background materials are presented in \Cref{tab:bg_materials_survey}. They were selected to offer a wide variety of prices and reflectivity values. These backgrounds are divided in three categories: fabric, paint and paper as offered by different manufacturers in the market. The values of their reflectivity suggest that the best commercially available option for space conditions recreation is the fabric Black Velvet, from KOYO, with the lowest reflectivity of all surveyed materials, 0.1\%. However, there are also materials available in the market whose reflectivity values are not readily available from manufacturers, making a direct comparison difficult.  Hence in \Cref{sec:backgroundtest} we propose an experimental analysis to compliment the market survey for evaluating the suitability of background materials to use in a space lab.

\begin{table}[!t]
\tablestyle[sansboldbw]
\centering
\caption{Survey of commercially available background materials selected to represent a different price and reflectivity options. }
\begin{center}
\footnotesize
\label{tab:bg_materials_survey}
\begin{tabular}{|m{1.35cm}|m{1.8cm}|m{1.7cm}|m{1.7cm}|m{1.5cm}|m{1.5cm}| m{1.8cm}|m{1.5cm}|m{1.4cm}|} 
\hline

\textbf{Item} & \textbf{Fineshut KIWAMY [1]} & \textbf{Fineshut SP [2]} & \textbf{Flock Sheet [3]} & \textbf{Black Velvet [4]} & \textbf{Neewer Background [5]} & \textbf{Musou Paint [6]} & \textbf{Black3.0 [7]} & \textbf{Background Paper [8]} \\ \hline

{Company} & KOYO & KOYO & KOYO & KOYO & Neewer & KOYO & Stuart Semple’s & Spectrum \\ \hline
{Material} & Fabric & Fabric & Fabric & Fabric & Fabric & Paint & Paint & Paper \\ \hline

{Composition} & fine urethane foam & fine urethane foam & rayon-base fabric electrostatically flocked with a  low-gloss nylon pile & rayon-base fabric / rayon pile, back coated with resin & - & synthetic resin (acrylic), pigment, antifungal agent, water & acrylic & - \\ \hline
{Reflectivity (\%)} & 0.75 & 1.25 & 1-5 & 0.1 & - & 0.6 & 2.5 & - \\ \hline

{Specs} & Thickness: 0.42±0.05mm; Standard sheet size: 53x280x480mm & Thickness: 0.22±0.03mm (Fineshut SP0.2); Standard size (roll): 500mmx130m &Thickness: 0.9±0.2mm; Dimensions: 950mmx20m (Max. size per roll) & Dimensions: 900mmx24m (max size per roll) & Dimensions: 3x3.6m & Lightfastness: ASTM Class II; A non-water-absorbent material may require a surface treatment or primer & Volume: 150ml,1L,6L & Badabing Black; Dimensions: 2.7 x10m \\ \hline

{Cost** (EUR)} & 88.18 for 480x280mm & 271.92 for 2500mmx10m & 227.82 for 950mmx20m & 455.64 for 900mmx10m & 53.99 & 52.74(100ml), 116.09(400ml), 263.90(1L) & 30.95(150ml), 120(1L), 643(6L) & 74.95 \\ \hline

\end{tabular}
\end{center}
\footnotesize
\raggedright
**As on June 2022 \\
\end{table}

\subsection{Cameras}
\label{sec:camera_survey}
For image capturing at a space lab facility, eight different cameras were surveyed. These cameras were selected to represent a wide cost range (from 25 to 3200 EUR) for cameras with different image capture capabilities. The results from the survey are presented in \Cref{tab:camera_survey}. However, the cameras need to be further compared in space-representative situations (varying illumination and exposure conditions). Hence, we propose additional experiments in \Cref{sec:exposuretest} for comparison of the cameras in such conditions. The market survey along with the experiments provided a comprehensive comparison of camera systems for a space lab. 

\begin{table}[!t]
\tablestyle[sansboldbw]
\centering
\caption{Survey of cameras available in the market to represent a wide price range and various exposure and resolution capabilities.}
\begin{center}
\footnotesize
\begin{tabular}{| m{1.5cm} | m{1.5cm}| m{1.5cm}| m{1.7cm}| m{1.5cm}| m{1.5cm}| m{1.5cm} |m{1.5cm} |m{1.5cm} |} 
 \hline
 \textbf{Camera} & \textbf{Sony A7RIII [9]} &\textbf{Canon EOS M200 [10]} & \textbf{Canon 5DSR [11]} & \textbf{iPhone 13 [12]} & \textbf{Intel RealSense D457 [13]} & \textbf{FLIR Blackfly S USB3 [14]} & \textbf{ Raspberry Pi (HQ) [15]} & \textbf{ Raspberry Pi (LQ) [16]}\\ [0.5ex] 
 \hline
 Focal length (mm) & 28-70 & 15-45 & 35 & 26 & 1.88  & - & - & 3.04\\ 
 
 \hline
 Shutter speed (s) & 1/8000 - 30 & 1/4000 -30 & 1/8000-30 &  1/8000 - 1/3 & 1/1000 -10 & 1/10\textsuperscript{6} - 30  & 1/8000$+$ & -\\
 \hline
 ISO range & 100-102400 & 10-25600 & 100-12800 & max. 7616 & - & - & 100 - 800 & 100-800\\
 \hline
 Maximum Resolution (MP) & 42.4 & 24.1 & 50.6 & 12 & 1 & 0.4 & 12.3 & 8\\
 \hline
 FPS & 1-100 & 23.98, 25 & 29.97,25,23.97  & max. 60 & 30  & max. 522  & max. 120 & max.90\\ [1ex] 
 \hline
Pixel size (um) & 4.51 & 3.72 & 4.14 & - & - & 6.9 & 1.55 & 1.4\\ [1ex] 
 \hline
 Weight (g) & 657 & 299 & 930 & 174  &  145 & 36 & - & 3\\ [1ex] 
 \hline
 Approx Price** (EUR) & 3200 & 550 & 1430 & 800  &  470 &  410 & 50 & 25\\ [1ex] 
 \hline
 Volume (cm\textsuperscript{3}) & 902 & 253 & 1351 &  80 & 129.456 & 25.23  & 25.5 & 6*\\ [1ex] 
 \hline
\end{tabular}
\end{center}
\footnotesize
\raggedright
HQ: High quality; LQ: Low quality \\
*Board size \\
**As on June 2022
\label{tab:camera_survey}
\end{table}

\subsection{Illumination lamps} 
\label{sec:lamps_survey}

\begin{table}[!t]
\tablestyle[sansboldbw]
\centering
\caption{Survey of illumination lamps selected to represent low, medium and high cost alternatives.}
\footnotesize
\begin{center} 
\begin{tabular}{|m{2.7cm} | m{2cm} |m{2cm}| m{2cm} |m{2cm} |m{2cm}  |m{2cm} |} 
 \hline
\textbf{Lamp} & \textbf{Small Reflectors [17]} & \textbf{Medium Reflectors [18]} & \textbf{Godox SL-60 [19]}  & \textbf{Low Glare Downlight [20]} & \textbf{Aputure LS 60d [21]} & \textbf{Sunbrick (sun simulator) [22]} \\ [0.5ex] 
 \hline
{Light source type} & COB LED & SMD LED  & COB LED & LED  & LED   & LED\\ 
 \hline
{Temperature / Wavelength}   & 2200-8000K & 3000-6000K & 5600$\pm$300K & 3000-5000K & 5600K  &  400-110nm\\ 
 \hline
{ Luminous flux (Lm)} & 136-1075 & 1650  & 4500 & - & 2715-54300  & 0.1-1.1 suns*\\ 
 \hline
{Power (W)} & max. 20 & 15  & 60-200 & 73-276 & 90 & 625\\ 
 \hline
{ Luminous efficiency (LPW)}& 130-160 & 110 & - &  121 & - & -\\ 
 \hline
{ Angle (deg)} & 16-36 & 78-83 & 70 & Omnidirectional & 15-45 & Omnidirectional \\
 \hline
{Dimensions (mm)} & D=35-75 H=29.4-75.5 & D=135-205 H=61.291.5 & 230x240x140 & 390x380x74, 480x380x74, 687x390x74 & 431.8 x 251.46 x 210.82 & 250x250x390 \\
 \hline
{Approx Cost** (EUR)} & 10-20 & 20-60 & 120 & 580-690 & 280 & 30810 \\
 \hline
{Brand} & Nata & Nata & Godox & Terralite Eco & Aputure & G2V Optics Inc. \\ [1ex] 
 \hline
\end{tabular}
\end{center}
\footnotesize
\raggedright
COB LED: Chip-on-Board, Light-Emitting-Diode; SMD LED: Surface-Mount-Device Light-Emitting-Diode \\ 
D: diameter;  H: height \\
*1-sun represents light that reproduces sunlight as specified in the ASTM E927 or IEC 60904-9 standards \\
**As on June 2022 \\
\label{tab:lamps_survey}
\end{table}

In \Cref{tab:lamps_survey}, a survey of commercially available illumination lamps is presented, selected to represent low, medium and high-cost alternatives. The lamps are assessed regarding their light source temperature/wavelength, luminous flux, power and efficiency, emission angle, total lamp dimensions, cost and brand. The low-cost alternatives include aluminum reflectors of small (10-20 EUR) and medium sizes (20-60 EUR) of around 15W that can be bought at regular home goods stores. 
A medium-cost alternative is proposed with an omnidirectional growth lamp (Low glare downlight, 580-690 EUR) with power up to 276W.
The most expensive alternative included in the survey is the large area solar simulator (Sunbrick, up to 30810 EUR), with programmable spectra and power up to 625W.

The section presented a detailed survey of different background materials, cameras and illumination lamps available in the market. However, this market survey alone is inadequate to make the equipment selection. For instance, the manufacturer may not always provide the needed metrics (as in the case with the background materials), or the equipment may need to operate in a different (harsh space-like) environment than its normal operating conditions (as for the cameras). Hence, to further reduce uncertainty, the survey is complemented with experimental analyses of the background materials and cameras. The laboratory setup for the experiments is described in the next section, followed by experimental analyses of background materials (\Cref{sec:backgroundtest}) and cameras (\Cref{sec:exposuretest}), respectively. The market survey, along with these supporting experimental analyses, will provide information to guide the selection of suitable image acquisition equipment in a space lab.

\section{Laboratory Setup}
\label{sec:labsetup_datacollection}

\begin{figure}[!t]
    \centering
    \includegraphics[scale=.7]{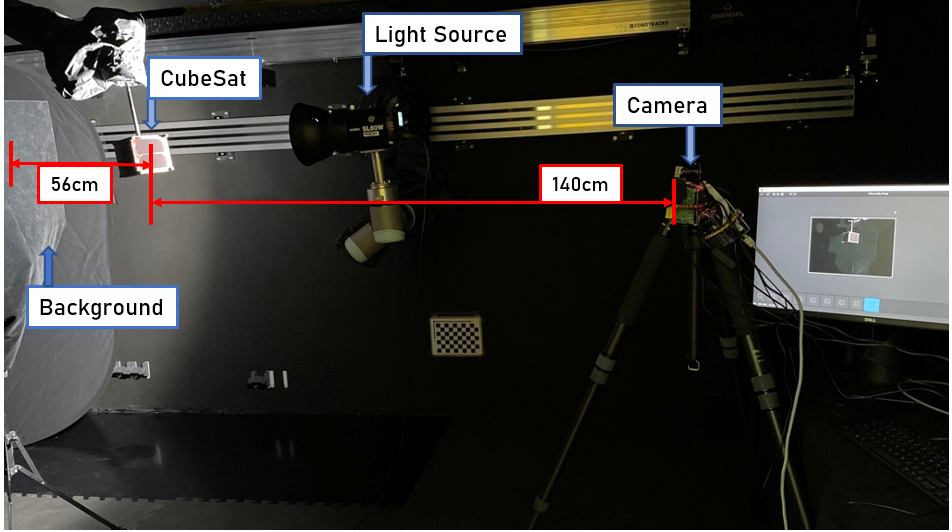}
    \caption{General laboratory setup for data collection using the SnT Zero-G Lab. The spacecraft (CubeSat, in this case) and the light source were mounted on the movable UR10e robotic arms, the cameras were mounted on a fixed tripod and the backgrounds were placed behind the CubeSat.}
    \label{fig:lab_setup}
\end{figure}

The data collection activities for the experiments presented in this article were conducted at the SnT Zero-G Lab facility. The SnT Zero-G Lab is a multipurpose facility capable of emulating a large variety of in-orbit operations in different orbital scenarios. The facility has two UR10e robotic arms mounted on rails, providing a 6+1 DoF. The robotic arms are capable of mimicking orbital trajectories of the spacecrafts, other orbital objects or the light sources. To recreate the challenging lighting conditions in space, the Zero-G Lab uses a  Godox SL-60 LED Video Light. The wall and ceiling are painted in black and epoxy flooring (black) is used to remove reflections and create a space like environment. 
The general setup of the laboratory environment for the experiments is shown in \Cref{fig:lab_setup} and consists of the following components:

\begin{itemize}
    \item \textbf{Cameras:} The cameras were mounted on a tripod directly facing the object of interest, i.e the spacecraft.
    \item \textbf{Spacecraft:} The spacecraft was mounted on a UR10e robotic arm. A 1U CubeSat was used in the experiments presented in this article. However, the experiments can be also be conducted using other types of spacecraft mock-ups available.
    \item \textbf{Background:} A dark background was placed behind the CubeSat, either mounted on a tripod-like structure or placed independently.
    \item \textbf{Light source:} A single continuous light source was mounted on a second UR10e robotic arm using a custom designed metal bracket fabricated in stainless steel.
 \end{itemize}
 
The experiment setup was designed taking into consideration constraints including the size of the room, visual light (VL) reflective surfaces present like the robotic rails, camera capability limitations and the physical restrictions related to the range of possible motion for the robotic arms. For all of the experiments, the camera position remains static while the robotic arms controled the light source and CubeSat positions. The trajectories/positions for the robotic arms were provided as a set of manually defined waypoints. Python scripts were used for automating the image capture process with different camera settings, to determine the appropriate white balance gain parameters and to avoid unwanted color shifts within images for each camera used. Black background materials were placed at a distance of $\sim$56cm to the rear of the vertical center of the CubeSat and the cameras were placed at a distance of $\sim$140cm  directly in front of the vertical center of the CubeSat.
 
\subsection{Cameras}
In the experiments, two different cameras were used: Raspberry Pi Low Quality (LQ) and High Quality (HQ) cameras. Technical specifications of the cameras are given in \Cref{tab:cameras}. These cameras were selected to represent both the low quality DIY cameras and high quality consumer-grade cameras.  The LQ camera relies on a inbuilt lens whereas the HQ cameras use a 12mm Edmond Optics lens. 

\begin{table}[!t]
\tablestyle[sansboldbw]
\centering
\caption{Technical specifications of the two cameras used in the data collection process}
\begin{tabular}{|p{4cm}|p{5cm}|p{5cm}|} 
\hline
\textbf{}  &   \textbf{LQ Camera}   & \textbf{HQ Camera} \\\hline
Camera	&   Raspberry Pi V2.0 & Rasberry Pi High Quality \\\hline 
Lens	&   Raspberry Pi 3.04mm & Edmunds Optical 12mm     \\\hline 
Focusing method &   Auto/Camera defined  & Manual \\\hline 
Image Size (h$\times$w) & 480$\times$640   &  480$\times$640  \\\hline 
Capture Format & jpeg + Bayer array   &  jpeg + Bayer array  \\\hline 
Auto White Balance &   Off  &   Off \\\hline 
Red Gain &  1.4883  &  3.1484  \\\hline
Blue Gain &    1.2539  &  1.5781    \\\hline 
Auto Exposure &   Off & Off\\\hline 
\end{tabular}
\label{tab:cameras}
\end{table}
 
\begin{figure}[!t]
	\centering
	\includegraphics[scale=.6]{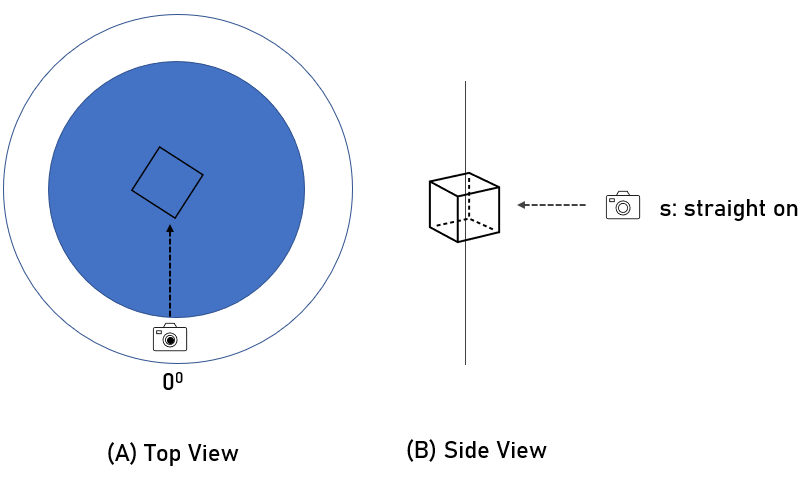}
	\caption{An illustration of the reference camera position, denoted as CP0, with respect to the CubeSat. (A) Top view and (B) Side view. Dotted arrows indicate the viewing direction.}
	\label{fig:CP0}
\end{figure}

For all of the experiments conducted, the reference camera position remained fixed and is denoted as CP0. In \Cref{fig:CP0}, CP0 is illustrated with the horizontal angle of 0\degree and the vertical position labelled \textbf{s}, indicating the camera was looking straight on the CubeSat at its central height.

\subsection{Lighting}
\label{sec:Lighting}
All experiments were conducted with a single light source (Godox SL-60 LED Video Light) available at the SnT Zero-G Lab facility. The technical specifications are given in \Cref{tab:lamps_survey}. Three lamp configurations were investigated, mimicking various illumination conditions from a space environment. For example, collimators, producing parallel light beams that create hard shadows and large differences in light intensity between illuminated
and dark regions, are typically chosen for mimicking objects in space illuminated by the sun without an atmosphere \citep{INVERITAS2015}. 
The lamp configurations used in the experiments are defined below:
\begin{itemize}
    \item \textbf{LAMP0:} Light source with a collimator as light modifier
    \item \textbf{LAMP1:} Light source with a reflector as light modifier
    \item \textbf{LAMP2:} Bare light source without any modifiers
\end{itemize}

Additionally, two different light intensities were also used for each of the lamp configurations: 
\begin{itemize}
    \item \textbf{Light Intensity Low (LIL):} 10$\%$ of the total light intensity available from the source.
    \item \textbf{Light Intensity High (LIH):} 100$\%$ of the total light intensity available from the source.
\end{itemize}

\begin{figure}[!t]
	\centering
	\includegraphics[scale=.5]{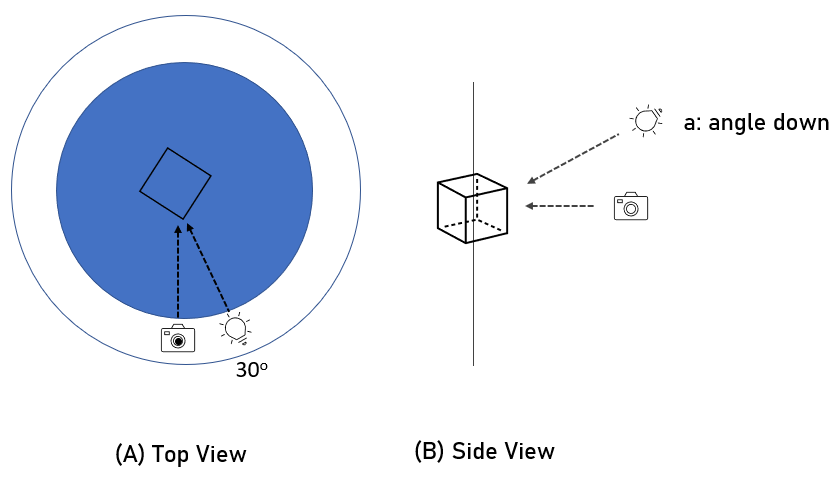}
	\caption{An illustration of the reference light position, denoted as LP0, with respect to the camera and the CubeSat. (A) Top view and (B) Side view. The dotted arrows indicate direction of viewing and lighting respectively.}
	\label{fig:LP0}
\end{figure}

The combination provided six (2 light intensities x 3 lamp configurations = 6) different illumination conditions used in the experiments. The reference lighting position (LP0) was defined as 30\degree  to the right of the camera in the vertical plane and above the camera height (a: angle down) horizontally,  as shown in \Cref{fig:LP0}. LP0 position enables the visualization of both shadows and highlights of the CubeSat and as such, best represented the 3D structure of the object when projected onto a 2D image plane. Details of other possible lighting positions, along with a qualitative comparison of the corresponding captured images are provided in supplementary material Section B. 

\section{Background Analysis}
\label{sec:backgroundtest}

The objective of the background materials experiment is to determine the background with the highest light absorption such that it appears featureless in the captured images. 
The backgrounds analysed in this experiment are: Black Velvet fabric (BG0), Moussu paint (BG1), Black 3.0 paint (BG2), Leitz-paper (BG3) and Neewer Background fabirc (BG4). 
Refer to \Cref{tab:bg_materials_survey} for more details.
In the context of the performed experiments, a ``featureless background'' corresponded to the one that added no discernible information to an image and most closely resembled the black colour, as represented by the RGB pixel value of (0,0,0).   

\subsection*{Data Collection}
\begin{figure}[!t]
	\centering
	\includegraphics[scale=.33]{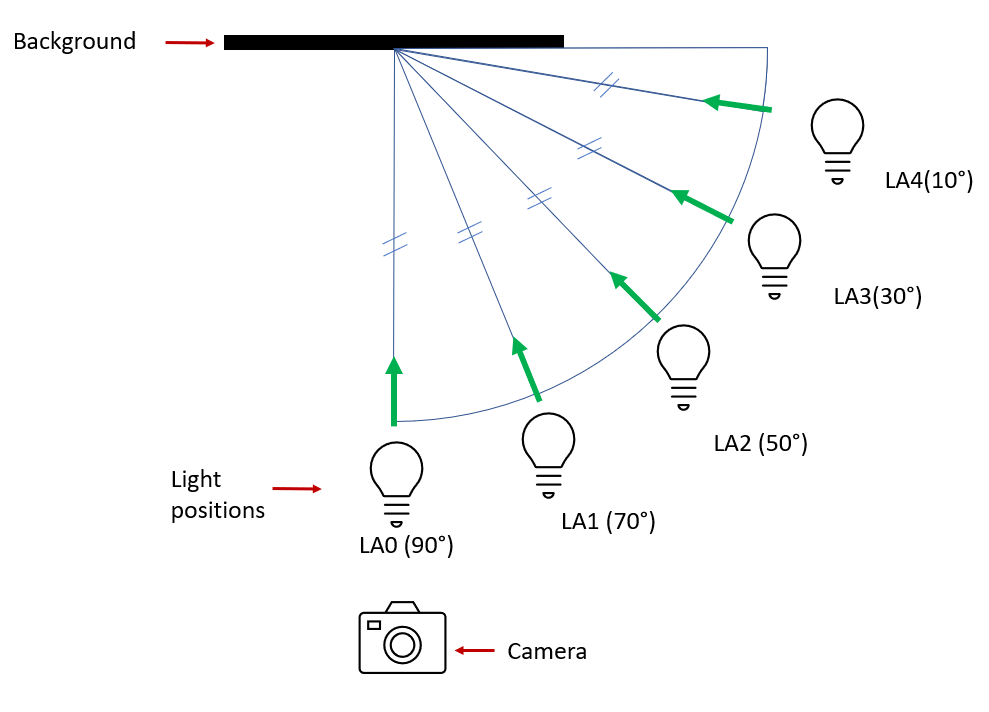}
	\caption{Background analysis data collection setup (top view).The green arrows indicates direction of lighting.}
	\label{fig:backgroundtestangles}
\end{figure}

\begin{figure}[!t]
	\centering
	\includegraphics[scale=.45]{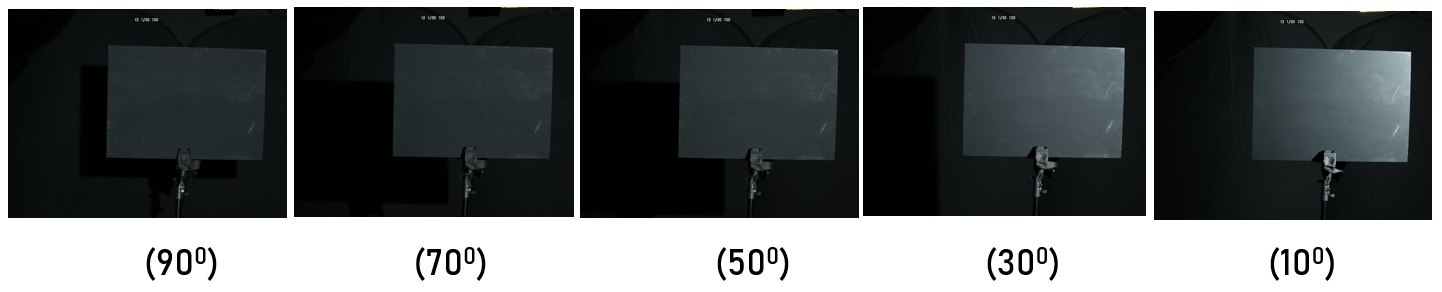}
	\caption{Sample images of BG2 collected for background analysis experiment. The images were captured with the HQ camera at the reference exposure with LAMP2 and the light intensity set to LIH. From top-left to bottom-right, the angles of incidence are: 90\degree, 70\degree, 50\degree, 30\degree, 10\degree.}
	\label{fig:bgsamples}
\end{figure}

For the data collection, the camera was set to position CP0 and the background to be tested was placed $\sim$196cm behind CP0. In this case, the CubeSat was not used and was, hence, removed from the camera's field of view. Images were captured for each of the background materials (BG0-4) under different light intensities (LIL and LIH), lamp configurations (LAMP0-2) and for five angles of illumination (LA0-4) as illustrated in \Cref{fig:backgroundtestangles}. A set of sample images collected are shown in \Cref{fig:bgsamples}. The captured images were cropped manually to include the background region before the experimental analysis.

\begin{figure}[!t]
	\centering
	\includegraphics[width=1\linewidth]{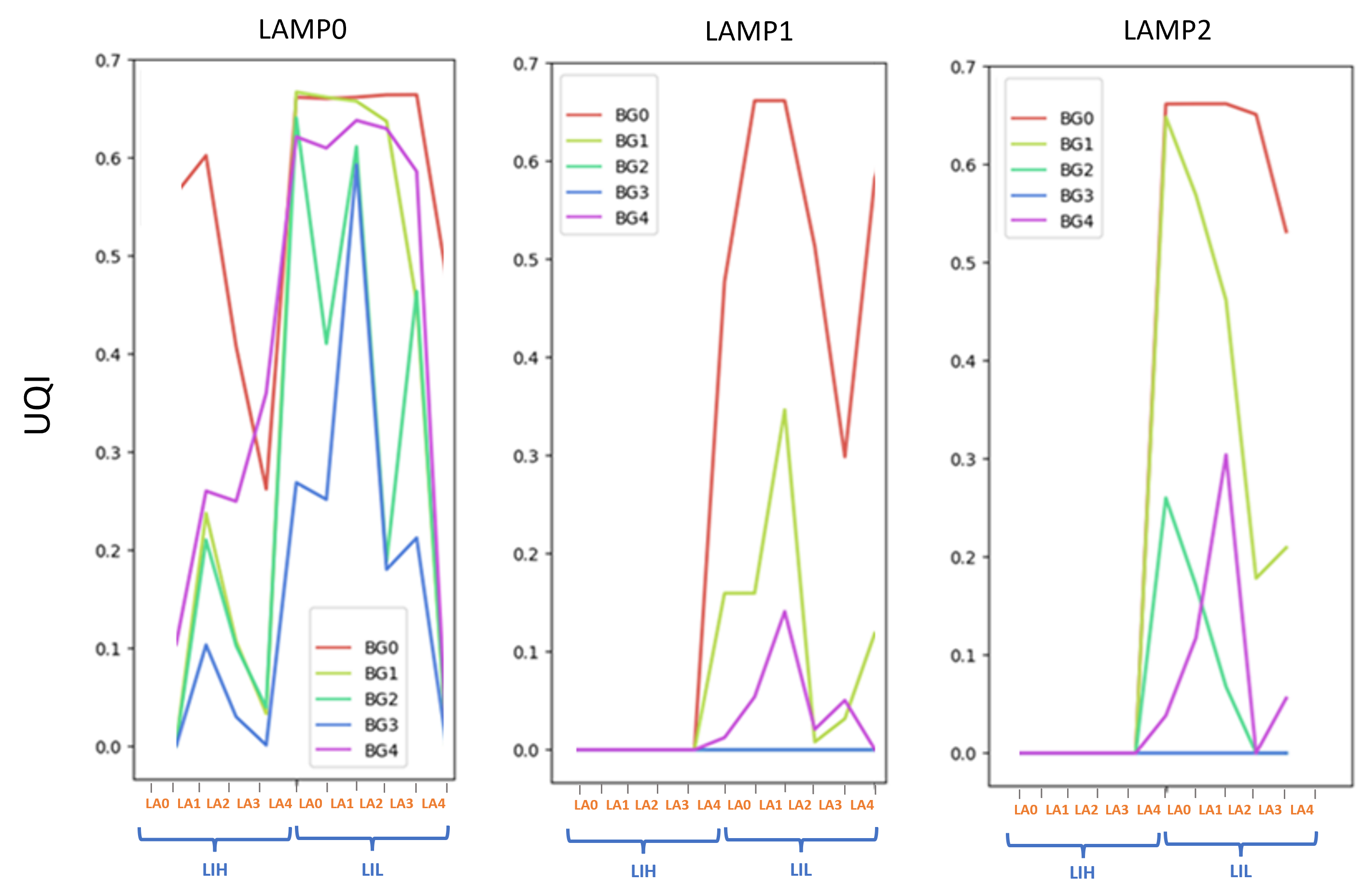}
	\caption{Background analysis results using LQ camera. The black velvet tissue (BG0) had the highest UQI values compared to all the other materials tested, and under different illumination conditions.}
	\label{fig:bgtestresults_lq}
\end{figure}

\begin{figure}[!t]
	\centering
	\includegraphics[width=.9\linewidth]{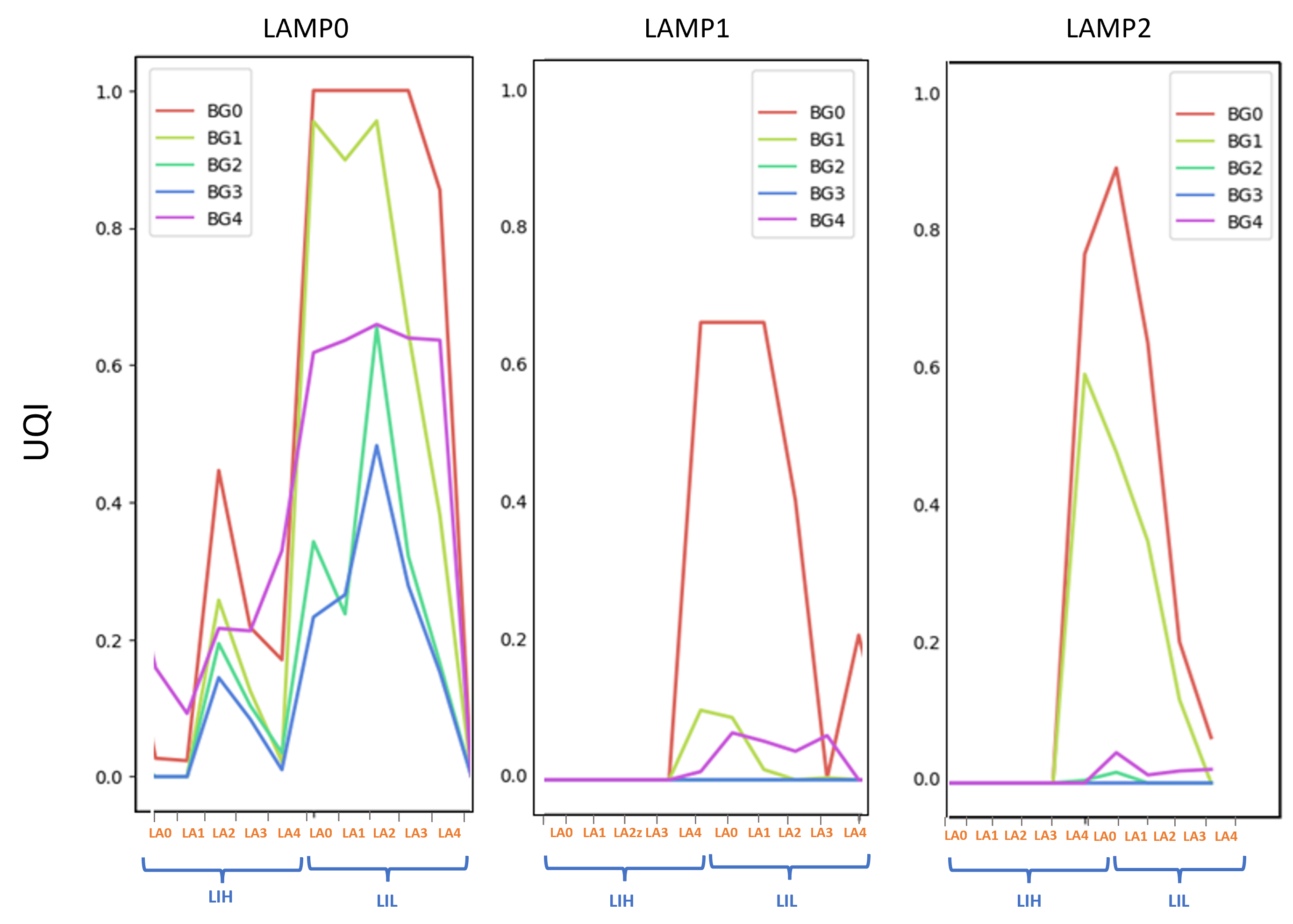}
	\caption{Background analysis results using HQ camera. The black velvet tissue (BG0) had the highest UQI values compared to all the other materials tested, and under different illumination conditions.}
	\label{fig:bgtestresults_hq}
\end{figure}

\subsection*{Experiment and results}
To evaluate the backgrounds, all the acquired images of each background were compared with a reference image. The chosen reference image was a pure black image synthetically generated by setting the RGB pixel values to (0,0,0). The comparison was quantified with an image similarity measuring index, the Universal Image Quality Index (UQI)~\citep{995823}. The UQI was preferred as it provides a significantly better comparison than the widely used distortion metric mean-squared error (MSE), due to its  well-defined mathematical framework.
Unlike the traditional error summation methods, UQI considers the following three factors for modelling any image distortion: loss of correlation, luminance distortion and contrast distortion. Given $\mathbf{x}$ and $\mathbf{y}$ as the input and reference image signals respectively, UQI can be formulated as:
\begin{equation}
UQI (\bar{x},\bar{y})=\frac{4 \sigma_{x y} \bar{x} \bar{y}}{\left(\sigma_{x}^{2}+\sigma_{y}^{2}\right)\left[(\bar{x})^{2}+(\bar{y})^{2}\right]},
\end{equation}
where $\bar{x}$, $\sigma_x$, $\bar{y}$ and $\sigma_y$ represent the mean and standard deviation of all the input and reference samples respectively, and $\sigma_{xy}$, the correlation. Moreover, UQI provides error measurements independent of the viewing conditions and individual observers (subjective analysis by humans). 

In \Cref{fig:bgtestresults_lq} and \Cref{fig:bgtestresults_hq} the UQI scores for all the acquired images (under different illumination conditions) with respect to the reference image for the LQ and HQ cameras respectively, are presented. It is evident that the black velvet fabric (BG0) had the highest UQI scores compared to all other backgrounds, both in the high intensity (LIH) and low intensity (LIL) light conditions. These results indicate that BG0 is the most featureless background with the highest light absorption, which is in agreement with the market survey (\Cref{sec:bg_survey}). This performance makes it the best choice, for image acquisition in a space lab. For the rest of the experiments detailed in this article, the BG0 was used.

\section{Camera Analysis}
\label{sec:exposuretest}

Experiments were performed with the two Raspberry Pi cameras (LQ and HQ) with the objective of performing a: 

\begin{itemize}
    \item Qualitative comparison of the ideally exposed images under varying illumination conditions for the LQ and HQ cameras.
    \item Quantitative study of the image quality degradation with different exposure settings (over-exposure and under-exposure) for both the cameras. 
\end{itemize}

Together, these analyses provided an experimental framework for comparing camera capabilities. Furthermore, along with the market survey presented in \Cref{sec:camera_survey}, it can support in making purchase choices for cameras selection in a space lab.

\subsection{Ideal Exposure Analysis}
\label{sec:ideal_exposure_exp}

Camera exposure settings are defined by three parameters: Aperture, Shutter Speed, and ISO (Gain), known as the exposure tuple (A:SS:ISO). The exposure tuple determine a given exposure value (EV), multiple exposure tuples can result in the same EV \citep{prakel2009basics}. Images captured with small values of aperture, for example, f/2, will allow more light to reach an image sensor for fixed shutter speed and ISO settings, which can be a useful property when capturing images in low light conditions. However, the choice of aperture also impacts the Depth Of Field (DOF) which determines which portions of a 3D object, relative to the focal plane,  will appear in focus when projected onto a 2D image plane. The smaller the value of aperture, the shallower the DOF. 
Therefore, a trade-off exists between image sharpness and brightness when an aperture setting is chosen. 
Another consideration is the focal length of the lens employed. The greater the lens focal length, the shallower the DOF for a fixed value of aperture. The shutter speed parameter dictates how long a camera's image sensor will receive light. Whether an image will be properly exposed is also a function of the shutter speed. The ISO parameter of the exposure setting impacts how sensitive an image sensor is to light. However, more noise will be introduced into an image if the ISO value is set to be more sensitive to light. In the performed experiments, to not introduce unwanted color shifts into images taken with the same camera, ``auto white balance'' was disabled and the Red and Blue Gain settings listed in Table \ref{tab:cameras} were applied for each camera.

\subsection*{Ideal exposure}  
A careful choice of the exposure tuple is required to generate an ``ideally exposed'' image within the context of the given illumination conditions and the mechanical limitations of the image capture device (camera). The concept of ideal exposure is application dependent. In the case of the performed experiments, ideally exposed images were those which had the CubeSat well illuminated with all the features (like edges, corners and surface panels) clear and distinguishable. 
To obtain the initial ideal exposure settings under different illumination conditions, a Sekonic L-558R DualMaster light meter was placed directly in front of the spacecraft object. Then, a careful visual inspection of images captured with further fine-tuned exposure settings was used to define the ideal exposure setting for the experiment. 

\subsection*{Data collection}
\label{exposure_exp1_data}
The cameras and the lamp were mounted at their reference positions CP0 and LP0, the  CubeSat was positioned between the background and camera at a distance of $\sim$56cm and $\sim$140cm respectively as shown in  \Cref{fig:lab_setup}. The images were captured under different light intensities (LIL and LIH) and lamp configurations (LAMP0-2).

\subsection*{Experiment and Results}

\begin{figure}[!t]
	\centering
	\includegraphics[scale=.55]{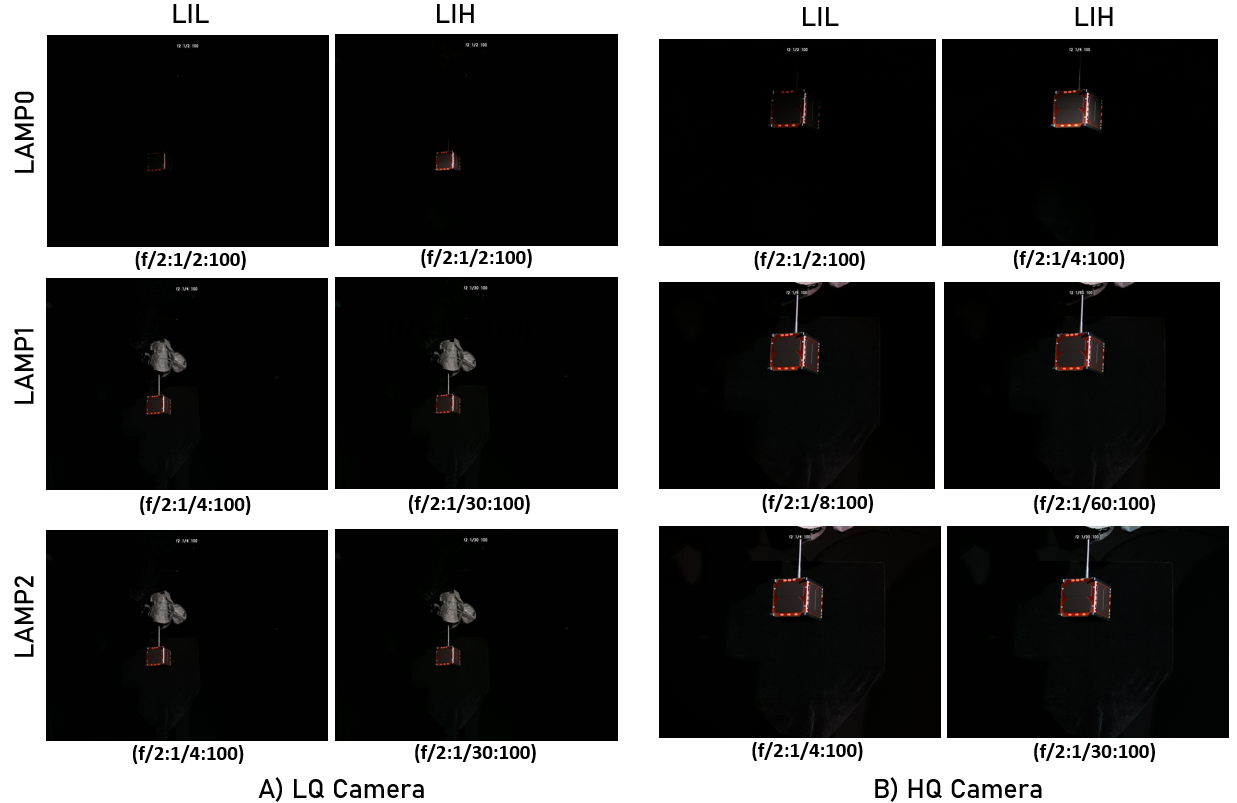}
	\caption{Ideally exposed images under different illumination conditions with corresponding exposure settings.The results suggest that careful adjustment of exposure settings can result in well exposed images with clear and distinguishable features for the object of interest.}
	\label{fig:ideal_exposure_images}
\end{figure}

Ideal exposure settings for images captured under different illumination conditions were obtained, as described with a light meter and by visual inspection. The selected ideal exposure settings with the corresponding images are shown in \Cref{fig:ideal_exposure_images}. The qualitative analysis of these images showed that, by carefully selecting the exposure settings, good quality well-exposed images can be obtained for different quality cameras under varying illumination conditions. This is particularly relevant in space-like environment where well exposed images with clear and distinguishable features need to be captured under considerable variations  in illumination conditions. The results also suggest that even with a LQ camera, images of good quality can be captured if exposure values are well calibrated.

\subsection{Exposure and Image Quality Analysis}
\label{sec:exposure_image_quality}

A reference exposure (EX0) was defined to study the effect of overexposure and underexposure in image quality. An aperture value of f/2.0 was selected as it allowed for an acceptable DOF and was achievable with all camera lenses tested. In addition, an aperture value of f/2.0 made it possible to capture images in a low-light setting at a shutter speed that would not introduce motion blur in the established laboratory setting. The reference shutter speed was set at 1/30\textsuperscript{th} of a second. Finally, the ISO value of 100 was chosen so as to not introduce unwanted noise into the captured images. Thus the reference exposure for capturing reference exposure images in this experiment is defined with the exposure tuple values of (f/2:1/30:100). 

The reference light intensity (LI0) corresponds to the light intensity required to set the reference exposure EX0 as the ideal exposure for each of the lamp configuration tested. To establish the LI0 light intensity, a light source was placed at the reference light position LP0 as shown in \Cref{fig:LP0}. The light intensity was then adjusted until the reference exposure provided ideal exposure. The LI0 values for LAMP0, LAMP1 and LAMP2 configurations were 75\%, 25\% and 30\% respectively. 

\subsection*{Data Collection}
Images were collected under the same positional setup described in \cref{exposure_exp1_data}. The images were captured only for the reference light intensities (LI0) and for lamp configurations (LAMP0-2) with each of the exposure settings defined in \Cref{tab:exposuresettings}. In this experiment, the underexposed and overexposed (EO1-EEO) conditions were obtained by changing only the camera shutter speed. The aperture and ISO values were kept the same. The HQ camera captured images with a 12mm lens, while the LQ camera had an inbuilt 6mm lens which effects the field of view in captured images. Also, because the cameras were mounted side-by-side, the cameras' view fields were slightly horizontally translated. Therefore, images were cropped to centrally align the CubeSat prior to analysis.

\begin{table}[!t]
\tablestyle[sansboldbw]
\centering
\caption{Details of different exposure settings used. EEU-EU1 denotes under-exposure, EX0 the reference exposure, and EO1-EEO are the over-exposure settings.}
\begin{tabular}{|l|c|c|c|c|}
\hline
\textbf{Exp. Setting Label}  &   \textbf{Aperture}   & \textbf{Shutter Speed (sec)}  &   \textbf{ISO}   & \textbf{EV} \\\hline 
EEU	&   f2  &   1/500   &   100 &   11  \\\hline 
EU3	&   f2  &   1/250	&   100 &   10  \\\hline 
EU2	&   f2  &   1/125	&   100 &   9   \\\hline 
EU1    &	f2  &   1/60	&   100 &  8 \\\hline 
EX0	    &   f2  &   1/30	&   100 &   7 \\\hline 
EO1	&   f2  &   1/15	&   100 &   6  \\\hline 
EO2	&   f2  &   1/8 	&   100 &   5   \\\hline 
EO3	&   f2  &   1/4 	&   100 &   4   \\\hline 
EEO	&   f2  &  1/2  	&   100 &   3   \\\hline 
\end{tabular}
\label{tab:exposuresettings}
\end{table}

\subsection*{Experiment and Results}
A quantitative analysis of the image quality degradation with varying exposure settings for a given illumination condition (light intensity and lamp configuration) was performed. Since the images contained a single CubeSat object, with a fixed background, the structural similarity with respect to the reference image provided a relevant measure of image quality degradation. 
The Multi-scale Structural Similarity Index (MS-SSIM)~\citep{1292216} was used to measure structural similarity. The MS-SSIM was derived from the Structural Similarity Index (SSIM)~\citep{wang2004image} extending it to incorporate multi-scale measures using image details at different resolutions. MS-SSIM separates the influence of illumination (average luminance and contrast) to explore the structural information in an image.

\begin{figure}[!t] 
	\centering
	\includegraphics[width=1\linewidth]{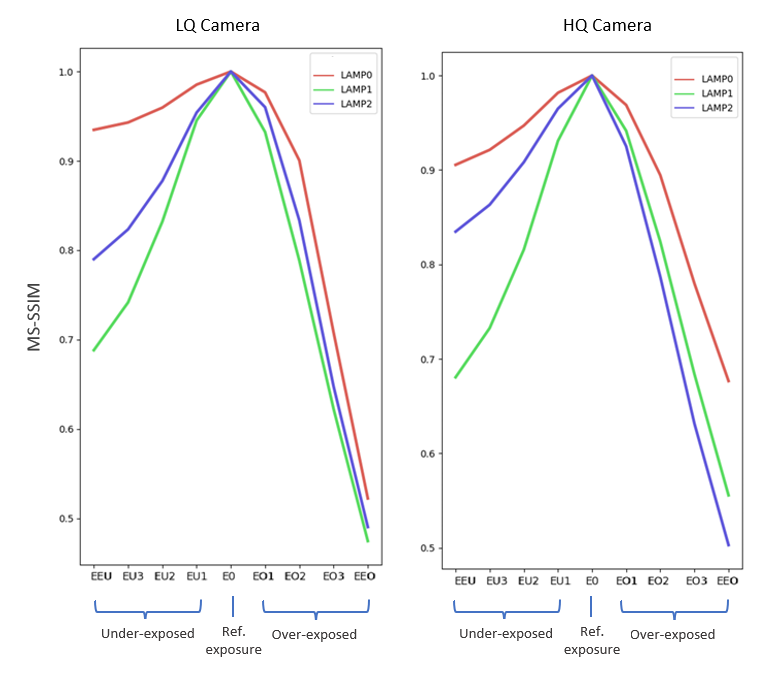}
	\caption{ Image degradation plots for LQ and HQ cameras with changes in exposure settings. EEU-EU1 are underexposed, E01-EE0 are overexposed and EX0 is the reference image. 
	The results suggest that both the cameras have similar relative image degradation under extreme exposure conditions.}
	\label{fig:exposuretestresults}
\end{figure}

 
In \Cref{fig:exposuretestresults}, the MS-SSIM score for the acquired images under different exposure settings with respect to the reference image, is presented. For the camera with a higher sensor quality (dynamic range), the image degradation was slower compared to one with a lower sensor quality. For both the cameras, the structural integrity of the images dropped identically on both sides of the curve (over and under exposed) with respect to the reference image. This behaviour indicates that the relative drop in image quality for both the cameras is similar under extreme exposure settings. 

\section{Discussion}
\label{sec:discussion}

The results of the performed experiments suggest that laboratory equipment selection is not a straightforward procedure where the most expensive options provide the best results. This was evidenced in the experiments performed with the Raspberry Pi cameras, where the most expensive cameras, those featuring the highest image resolutions for example, are not always significantly better than cheaper alternatives.  
As shown in \Cref{sec:ideal_exposure_exp}, a careful calibration of exposure settings can produce good quality ideally exposed images for both the LQ and HQ cameras. Similarly, \Cref{sec:exposure_image_quality} indicated that the relative image degradation in extreme exposure settings for both the cameras is identical; hence, the camera selection is highly application dependent and needs to be analysed experimentally case by case. Market surveys, such as the one presented in \Cref{sec:camera_survey} will serve as a start point and need to be followed by experimental analyses. 
The same criteria apply to the selection of the background material. The experimental results indicated that the highest UQI were obtained with black velvet fabric, which was an expected result as this was the material with the lowest VL reflectivity (\Cref{sec:bg_survey}). However, as different manufacturers might implement different methods to determine the reflectivity of their products (or might not even provide a reflectivity value at all), if possible, UQI (or even VL reflectivity) of dark background materials should also be tested before making a major purchase.
The SnT Zero-G Lab facility is still under development, further experiments will be conducted for camera analysis with images captured in scenarios where the space object moves along a trajectory relative to the camera. The objective of these experiments would be to introduce motion blur and other effects common during applications like vision-based navigation. 
Future work will also focus on conducting an experimental analysis of different lighting sources to support the market survey in this article. Spectral analysis will provide a more accurate comparison of light sources and help to analyse their similarity to real space conditions.

\section{Conclusion}
\label{sec:conclusion}
Facilities simulating real-world space environments are an integral part of training and validating vision-based space applications. High fidelity space-like images with annotations can be collected from these facilities to train and test the algorithms. However, development of such a space lab is challenging. The current literature lacks support in the form of manuals or templates. In this context, this article focused on a key aspect of a space lab facility development, the equipment selection. 
This article presented a systematic approach to equipment selection for image acquisition process, based on the lessons learned during SnT Zero-G Lab development at University of Luxembourg. The approach combines a market survey of equipment followed by experimental analysis. Background materials, cameras and the illumination lamps were surveyed. The background materials were first compared based on the VL reflectivity values obtained from the  manufacturers. 
For comparing materials with unknown reflectivity values, we present an experimental analysis method that calculates UQI scores with reference to a synthetically generated black image to identify suitable options. 
For camera selection, experiments suggest that a market survey alone will not provide sufficient information to make a purchase decision.
The results demonstrate that it is possible to obtain comparably good quality images even from a lower quality (less expensive) camera by carefully calibrating the exposure settings. Hence, for selecting camera systems, the market survey and experimental analysis should be used in tandem to gather the required information.
Future work is planned for studying image quality when motion blur and other phenomena are introduced, and for studying the performance of different light sources for simulating the space environment.

\putbib
\end{bibunit}

\clearpage
\begin{bibunit}
\renewcommand\thesection{\Alph{section}}
{{\Huge  Supplementary Material}}
\setcounter{section}{0}
\setcounter{figure}{0}
\setcounter{table}{0}

\section{Market Survey: Additional Information}
The reference links for each item surveyed in the article is given in \cref{tab:purchaselinks}. 

\begin{table}[!h]
\tablestyle[sansboldbw]
\centering
\caption{The reference links for each item surveyed}
\label{tab:purchaselinks}
\resizebox{\columnwidth}{!}{%
\begin{tabular}{|lll|}
\hline   
\multicolumn{1}{|l|}{\textbf{Ref}} & \multicolumn{1}{l|}{\textbf{Product}} & \textbf{Reference} \\ \hline
\multicolumn{3}{|c|}{\textbf{Background Materials}} \\ \hline
\multicolumn{1}{|l|}{\textbf{{[}1{]}}} & \multicolumn{1}{l|}{\begin{tabular}[c]{@{}l@{}}Fineshut KIWAMY\end{tabular}} & https://www.ko-pro.black/product/fs-kiwami/ \\ \hline
\multicolumn{1}{|l|}{\textbf{{[}2{]}}} & \multicolumn{1}{l|}{Fineshut SP} & https://www.ko-pro.black/product/fs-sp/ \\ \hline
\multicolumn{1}{|l|}{\textbf{{[}3{]}}} & \multicolumn{1}{l|}{Flock Sheet} & https://www.ko-pro.black/flocksheet/ \\ \hline
\multicolumn{1}{|l|}{\textbf{{[}4{]}}} & \multicolumn{1}{l|}{Black Velvet} & https://www.ko-pro.black/product/musou-black-fabric-kiwami/ \\ \hline
\multicolumn{1}{|l|}{\textbf{{[}5{]}}} & \multicolumn{1}{l|}{\begin{tabular}[c]{@{}l@{}}Neewer Background\end{tabular}} & https://www.amazon.com/Neewer-Collapsible-Background-Photography-Televison/dp/B00SR28SJ8 \\ \hline
\multicolumn{1}{|l|}{\textbf{{[}6{]}}} & \multicolumn{1}{l|}{Musou Paint} & https://www.ko-pro.black/product/musou-black-paint/ \\ \hline
\multicolumn{1}{|l|}{\textbf{{[}7{]}}} & \multicolumn{1}{l|}{Black3.0} & https://culturehustle.com/collections/black/products/black-3-0-the-worlds-blackest-black-acrylic-paint-150ml \\ \hline
\multicolumn{1}{|l|}{\textbf{{[}8{]}}} & \multicolumn{1}{l|}{\begin{tabular}[c]{@{}l@{}}Background Paper\end{tabular}} &  https://spectrum-brand.com/products/spectrum-badabing-black-non-reflective-paper-roll-backdrop-2-7-x-10m\\ \hline
\multicolumn{3}{|c|}{\textbf{Camera}} \\ \hline
\multicolumn{1}{|l|}{\textbf{{[}9{]}}} & \multicolumn{1}{l|}{Sony A7RIII} & https://www.amazon.com/Sony-a7R-Mirrorless-Camera-Interchangeable/dp/B076TGDHPT \\ \hline
\multicolumn{1}{|l|}{\textbf{{[}10{]}}} & \multicolumn{1}{l|}{Canon EOS M200} & https://www.canon.de/cameras/eos-m200/ \\ \hline
\multicolumn{1}{|l|}{\textbf{{[}11{]}}} & \multicolumn{1}{l|}{Canon 5DSR} & https://www.bhphotovideo.com/c/product/1119027-REG/canon\_0582c002\_eos\_5ds\_r\_dslr.html \\ \hline
\multicolumn{1}{|l|}{\textbf{{[}12{]}}} & \multicolumn{1}{l|}{\begin{tabular}[c]{@{}l@{}}iPhone 13\end{tabular}} &  https://www.apple.com/iphone-13/key-features/ \\ \hline
\multicolumn{1}{|l|}{\textbf{{[}13{]}}} & \multicolumn{1}{l|}{Intel RealSense D457} & https://www.intelrealsense.com/depth-camera-d457/ \\ \hline 
\multicolumn{1}{|l|}{\textbf{{[}14{]}}} & \multicolumn{1}{l|}{Blackfly S USB3
} & https://www.flir.eu/products/blackfly-s-usb3/ \\ \hline
\multicolumn{1}{|l|}{\textbf{{[}15{]}}} & \multicolumn{1}{l|}{\begin{tabular}[c]{@{}l@{}}Raspberry Pi (HQ)\end{tabular}} &  https://www.electronic-shop.lu/product/182978?src=raspberrypi \\ \hline
\multicolumn{1}{|l|}{\textbf{{[}16{]}}} & \multicolumn{1}{l|}{\begin{tabular}[c]{@{}l@{}}Raspberry Pi (LQ)\end{tabular}} & https://www.raspberrypi.com/products/camera-module-v2/ \\ \hline
\multicolumn{3}{|c|}{\textbf{Illumination Lamps}} \\ \hline
\multicolumn{1}{|l|}{\textbf{{[}17{]}}} & \multicolumn{1}{l|}{Small
Reflectors} &  https://www.nata.cn/images/download/nata\_catalog.pdf \\ \hline
\multicolumn{1}{|l|}{\textbf{{[}18{]}}} & \multicolumn{1}{l|}{Medium
Reflectors} & https://www.nata.cn/images/download/nata\_catalog.pdf \\ \hline
\multicolumn{1}{|l|}{\textbf{{[}19{]}}} & \multicolumn{1}{l|}{Godox SL-60} & https://www.bhphotovideo.com/c/product/1341997-REG/godox\_sl60w\_5600k\_60w\_white.html \\ \hline
\multicolumn{1}{|l|}{\textbf{{[}20{]}}} & \multicolumn{1}{l|}{Low Glare
Downlight} & https://img.roline.ch/publikationen/grah2019\_news.pdf \\ \hline
\multicolumn{1}{|l|}{\textbf{{[}21{]}}} & \multicolumn{1}{l|}{\begin{tabular}[c]{@{}l@{}}Apurture LS 60d\end{tabular}} & https://www.aputure.com/products/ls-60d/ \\ \hline
\multicolumn{1}{|l|}{\textbf{{[}22{]}}} & \multicolumn{1}{l|}{\begin{tabular}[c]{@{}l@{}}Sunbrick (sun simulator)\end{tabular}} & https://g2voptics.com/products/sunbrick-solar-simulator/ \\ \hline
\end{tabular}%
}
\end{table}

\section{Comparison of Lighting Positions}
\label{sec:lightingtest}

The objective of the analysis is to show the impact of different lighting positions on 3D objects projected onto 2D image planes that emulate those most often encountered in space environments. 
 
\subsection{Data Collection} 
\begin{figure}[t]
	\centering
	\includegraphics[width=.8\linewidth]{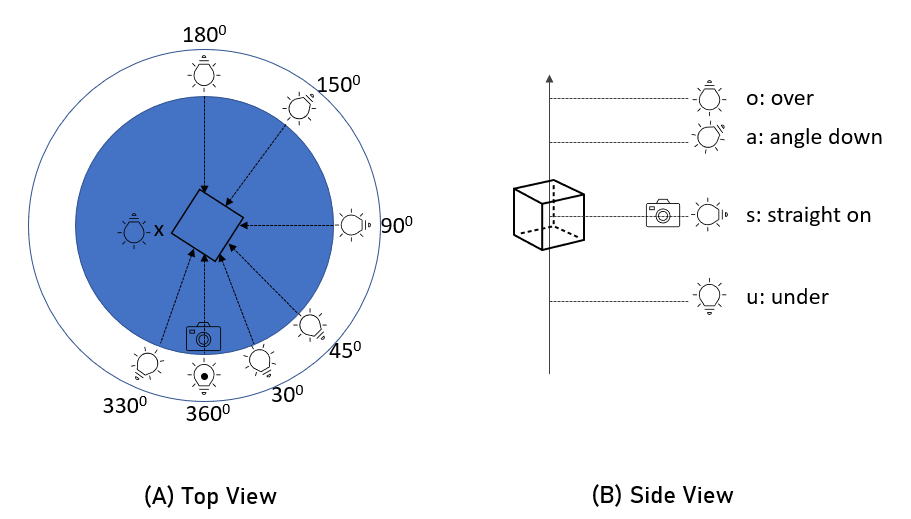}
	\caption{Camera and lighting position  with respect to the spacecraft object for the comparison of lighting positions}
	\label{fig:lightingsettings}
\end{figure}

The camera is set to position CP0. The background (BG0) is positioned 196cm away from the camera and the CubeSat is positioned in between them (56cm away from the background and 140cm away from the camera). The lighting positions LP0 through LP8 are defined in \Cref{tab:LP} and illustrated in \Cref{fig:lightingsettings}. The light setting descriptions are loosely related to terms often used in professional photography and show a wide range of possible illumination conditions \citep{LightScienceMagic}. Due to the movement restrictions of the UR10 robotic arm on which the light source is mounted, the LP5 (Broad lighting) is achieved by setting the light to position LP0 and rotating the CubeSat to simulate the desired lighting effect. Images are captured under the three light intensities (LIL, LIH), lamp configurations (LAMP0-2) and for each of the lighting positions (LI0-8).

\begin{table}[!t]
\tablestyle[sansboldbw]
\centering
\caption{Lighting position labels, angles, height and description for the comparison of lighting positions. *LP5 is achieved by setting the lighting to position LP0 and rotating the CubeSat to simulate the desired lighting effect.** Denotes origin, i.e. where the CubeSat is placed.}
\begin{tabular}{|l|r|c|l|}
\hline
\textbf{Lighting Position Label}  &   \textbf{Angle}   & \textbf{Height}  &   \textbf{Description}\\\hline 
LP0	&   30  &   a   &	Loop        \\  \hline
LP1	&   45  &   a	&  Rembrandt    \\  \hline
LP2	&   90  &   a	&   Side        \\  \hline
LP3 &	150 &   a	&   Rim         \\  \hline
LP4	&   180 &   a	&   Back        \\  \hline
LP5*	&   330 &   a	&   Broad       \\  \hline
LP6	&   360 &   a	&   Front       \\  \hline
LP7	&   **x   &   o	&   Top         \\  \hline
LP8	&   **x   &   u	&   Under       \\  \hline
\end{tabular}
\label{tab:LP}
\end{table}

\begin{figure}[!t]
	\centering
	\includegraphics[scale=.8]{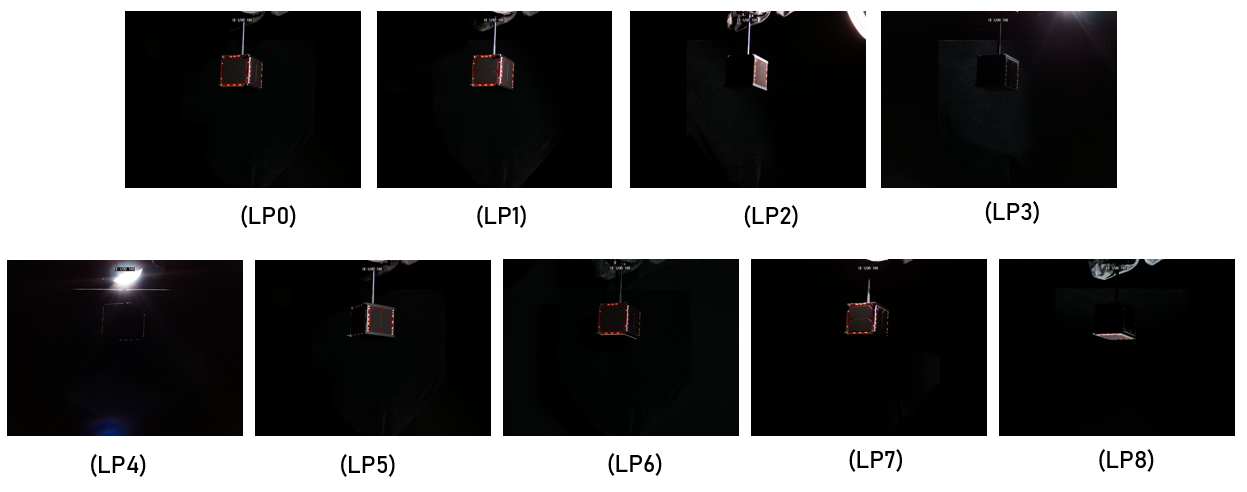}
	\caption{Sample images collected for lighting tests. The shown images are captured with the HQ camera at the LI0 light intensity setting for LAMP1 configuration.}
	\label{fig:lighttestsamples}
\end{figure}

\subsection{Discussion} 
A set of sample images captured for positions LP0 to LP8 in show in \Cref{fig:lighttestsamples}. The results show LP0 position to be the ideal lighting position for capturing images of highest quality. LP0 position shows both shadows and highlights of the CubeSat and as such, best represents the 3D structure of the object when projected onto a 2D image plane. 

\putbib
\end{bibunit}


\end{document}